\pdfoutput=1

\documentclass[11pt]{article}

\usepackage[]{acl_latex}

\usepackage{times}
\usepackage{latexsym}

\usepackage[T1]{fontenc}

\usepackage[utf8]{inputenc}

\usepackage{tabularx}
\usepackage{graphicx}
\usepackage{array}
\usepackage{float}
\restylefloat{table}
\usepackage{adjustbox}
\usepackage{array, multirow}
\usepackage{comment}
\usepackage{xcolor}
\usepackage{amsmath}
\usepackage{bm}

\usepackage[english]{babel}
\usepackage{blindtext}
\usepackage{bbm}

\usepackage{color}
\usepackage{tcolorbox}
\usepackage{CJK}
\usepackage{adjustbox}
\usepackage{booktabs}
\tcbset{width=0.4\textwidth,boxrule=0pt,colback=red,arc=0pt,auto outer arc,left=0pt,right=0pt,boxsep=5pt}
\usepackage{cleveref}

\usepackage{microtype}

\title{Domain Classification-based Source-specific Term Penalization for\\ Domain Adaptation in Hate-speech Detection}

\author{Tulika Bose \textsuperscript{$\dagger$} \quad Nikolaos Aletras \textsuperscript{$\ddagger$} \quad Irina Illina \textsuperscript{$\dagger$} \quad Dominique Fohr \textsuperscript{$\dagger$}  \\
  \textsuperscript{$\dagger$} Universite de Lorraine, CNRS, Inria, LORIA, F-54000 Nancy, France\\
  \textsuperscript{$\ddagger$}University of Sheffield, United Kingdom\\
  {\tt \{tulika.bose, illina, dominique.fohr\}@loria.fr}\\ 
  {\tt n.aletras@sheffield.ac.uk}
}

\begin{document}
\maketitle
\begin{abstract}
\emph{{\bf Warning: } this paper contains content that may be offensive and distressing.}

State-of-the-art approaches for hate-speech detection usually exhibit poor
performance in out-of-domain settings.
This occurs, typically, due to classifiers overemphasizing source-specific information that negatively impacts its domain invariance. Prior work has attempted to penalize terms related to hate-speech from manually curated lists using feature attribution methods, which quantify the importance assigned to input terms by the classifier when making a prediction. 
We, instead, propose a domain adaptation approach that automatically extracts and penalizes source-specific terms using a domain classifier, which learns to differentiate between domains, and feature-attribution scores for hate-speech classes,  yielding consistent improvements in cross-domain evaluation.
\end{abstract}

\section{Introduction}

While recent state-of-the-art hate-speech classifiers~\citep{AYO2021114762, dsa-etal-2020-towards,mozafari2019bert} yield impressive performance on in-domain held-out instances,
they suffer when evaluated on out-of-domain settings
\citep{yin2021generalisable,10.1145/3331184.3331262,swamy-etal-2019-studying,karan-snajder-2018-cross}. The distributions across corpora$/$domains\footnote{We use the terms `corpus' and `domain' interchangeably.} change due to varying vocabulary, topics of discussion over time~\citep{app10124180,10.1145/2124295.2124376}, data bias caused by sampling strategies \citep{wiegand-etal-2019-detection} and different hate-targets. This is concerning since curating new data resources for hate-speech involves substantial time and effort \citep{Poletto2019AnnotatingHS,malmasi:2018:profanity}. 
This calls for strategies, like Domain Adaptation (DA) approaches, that can adapt models trained on existing labeled resources to a new target domain that lacks class-labels.  

However, research on
DA in hate-speech is limited \citep{Sarwar2022UnsupervisedDA,bashar2021progressive,bose-etal-2021-unsupervised}. 
Typically, vanilla classifiers tend to learn more from domain-specific features \citep{YE202161,wiegand-etal-2019-detection} than domain-invariant features, resulting in poor 
out-of-domain performance.
For instance,
\citet{wiegand-etal-2019-detection} show 
that in a hate-speech dataset \citep{waseem},  neutral domain-specific terms, like `\textit{football}', `\textit{commentator}', etc., discussing the role of women in sports, are highly correlated with the hate label, restricting its generalizability.  
Thus, it is worth minimizing the importance of such terms
for improving cross-domain performance.

Recently, feature attributions -- 
methods for extracting 
post-hoc model explanations,  
have been used to align features with prior domain knowledge \citep{pmlr-v119-rieger20a, 
NEURIPS2020_075b051e}. These provide importance scores to the input terms as per their contribution towards the model prediction \citep{NIPS2017_8a20a862}. For instance, \citet{liu-avci-2019-incorporating,kennedy-etal-2020-contextualizing} reduce the over-sensitivity of classifiers on a curated list of 
identity terms
(e.g. \textit{muslims}, \textit{gay}) by penalizing their importance.
However, newly emerging social-media terms \citep{Grieve2018MappingLI} may render such lists non-exhaustive. \citet{yao2021refining} do not use any list but they require human-provided refinement advice as inputs. \citet{chrysostomou2022empirical} further show that post-hoc explanation methods might not provide faithful explanations in out-of-domain settings. The contemporaneous work by \citet{attanasio-2022-entropy} and \citet{D-Ref} reduce lexical overfitting automatically with 
entropy-based attentions and feature attributions, respectively.
While cross-domain classification performance across different datasets is not studied in the former, the latter needs some labeled target instances to identify the over-fitted terms.

In the task of detecting objects in images, \citet{zunino2021explainable} use a domain classifier, trained to differentiate between domains, to visually identify the irrelevant background information  
to be domain-specific.
Thus, they enforce the model explanations to align with the ground-truth annotations 
highlighting the objects in the image.
Inspired by this, we propose a new DA approach in hate-speech employing a domain classifier, but without having access to such annotations for aligning the attribution scores. 

We hypothesize that domain-specific terms that are simultaneously predictive of the hate-speech labels are instrumental in restricting the domain invariance of the hate-speech classifier. 
To this end, we employ a domain classifier to automatically extract the terms that help in identifying the source domain compared to the \textit{unlabeled}  target domain,
and feature-attribution scores to identify the subset important for hate-speech classification from the source.
\textit{Our method, through penalization of these terms, automatically enforces the source domain classifier to focus on domain-invariant content. }
Compared to approaches transforming high-dimensional intermediate representations to reduce the domain discrepancy, such as domain adversarial learning \cite{Ryu2020KnowledgeDF,tzeng2017adversarial}, our approach makes the adaptation more explainable,
while improving the overall cross-domain performance compared to prior-approaches.

\section{Proposed Approach}

Given training data from a labeled source domain  $D_S^{train}$ and an unlabeled target domain  $D_T^{train}$, 
our approach for DA in hate-speech involves 2 steps: (i) extraction of source-specific terms and (ii) reducing the importance of these terms. Our setting is similar to \citet{ben-david-etal-2020-perl} and \citet{Ryu2020KnowledgeDF}.

\subsection{Extraction of Source-specific Terms}

\paragraph{Domain classification} To identify source-specific terms, we first train a binary domain classifier using $D_S^{train}$ and $D_T^{train}$ that learns to identify whether a candidate instance comes from the source or the target domain. For this, we use a simple Logistic Regression (LR) with bag-of-words, as it is inherently interpretable.
We then use its feature weights to extract the top $N$ most important terms for predicting the source domain class. Each term is tokenized with the BERT \citep{devlin-etal-2019-bert} WordPiece tokenizer for compatibility with transformer models. 
The top $N$
terms obtained through domain classification are denoted as $S_{LR}$. 

\paragraph{Attribution-based term ranking} Intuitively, the terms 
from $S_{LR}$ that also contribute highly to the hate-speech labels, are likely to restrict generalization to the target as they could potentially reduce the importance assigned by the classifier to domain-invariant hate-speech terms. Thus,
we extract only those source-specific terms that are highly correlated with the labels, given the binary classification task of \textit{hate} versus \textit{non-hate}.

To this end, we first continue pre-training BERT on the unlabeled $D_T^{train}$ using the Masked Language Model (MLM) objective for incorporating the language-variations of the target domain, following \citet{glavas-etal-2020-xhate}. We then perform supervised classification on $D_S^{train}$ using this MLM trained model.
After every epoch, 
we obtain 2 ranked lists of terms for the two classes, sorted in the
order of decreasing importance.
We construct the lists using feature attribution methods that yield 
instance-level attribution scores $\text{ins\mbox{-}atr}^j_{te}$ per term $te$ in an instance $j$ -- a higher score indicating a higher contribution to the predicted class. We discard the scores of stop-words and the infrequent terms, and normalize $\text{ins\mbox{-}atr}^j_{te}$ using the sigmoid function. For obtaining a corpus-level class-specific attribution score $\text{cp\mbox{-}atr}_{te}^c$ per term $te$ and per class $c$, we perform a corpus-level average of all the $\text{ins\mbox{-}atr}^j_{te}$ for every $c$ using Equation \ref{eq1}. 
\begin{equation}\label{eq1}
  \resizebox{0.85\columnwidth}{!}{%
  $\text{cp\mbox{-}atr}_{te}^c = \frac{\sum_{j=1}^{|D_S^{train}|}\mathbbm{1}_{\hat{y_j}=c}\text{ins\mbox{-}atr}_{te}^j \forall  \text{occurrence of $te$ in j}}{\sum_{j=1}^{|D_S^{train}|}\mathbbm{1}_{\hat{y_j} = c} \#(\text{occurrence of $te$ in j})}$%
  }
\end{equation}
Here $c$ $\in$ \{hate, non-hate\}, $\hat{y}$ is the predicted class and $\mathbbm{1}$ is the indicator function. We sort the scores $\text{cp\mbox{-}atr}_{te}^c$ for all $te$ to obtain the highest attributed (i.e. most important) term per class to the lowest, yielding the ranked lists of terms per class, given by $\text{CP} = [{cp\mbox{-}hate}, {cp\mbox{-}non\mbox{-}hate}]$. 

We extract the source-specific terms $\text{te}^S$
that are common to both $S_{LR}$ and the top $M$ terms from $\text{CP}$, i.e. $\text{te}^S = [te \in S_{LR} \And {te} \in top_M(\text{CP})]$. These steps are repeated after every epoch. 
Note that the list $S_{LR}$ remains constant across the epochs, as it is independent to the hate-speech classification task.

\subsection{Penalization of Source-specific Terms}
We hypothesize that
penalizing $\text{te}^S$ obtained from the previous epoch during the next epoch 
should reduce the importance of terms that are both (i) domain-specific and (ii) contribute highly to the source labels, and thus, help learn from domain invariant terms.
We minimize the attribution scores for $\text{te}^S$, with $L_2$ penalization, in Equation \ref{eq2}.
\begin{equation}\label{eq2}
\resizebox{0.85\hsize}{!}{%
    $\mathcal{L} = \mathcal{L^{'}} + \lambda \mathcal{L}_{\textrm{atr}};  \mathcal{L}_\textrm{atr} = \sum \limits_{\text{t}\in \text{te}^S} \phi \left(\text{t}\right)^\textsuperscript{2}; \text{t} \in \text{te}^S$%
    }
\end{equation}
Here $\mathcal{L^{'}}$ is the 
classification loss and $\mathcal{L_{\text{atr}}}$
is the attribution loss. $\lambda$ controls the strength of penalization, and $\phi \left(\text{t}\right)$ is the attribution score 
for t.

We experiment with two variations: (i) \textbf{Dom-spec:} penalizing only the terms in $\text{te}^S$; (ii) \textbf{Comb}: penalizing the combination of $\text{te}^S$ and the terms from \citet{liu-avci-2019-incorporating,kennedy-etal-2020-contextualizing}.

We use two different feature attribution methods that have been widely used in recent studies~\citep{chrysostomou-aletras-2021-improving,chrysostomou2022flexible}: (i) \textbf{Scaled Attention ($\alpha \nabla \alpha$}) \cite{serrano-smith-2019-attention}, which scales attention weights $\alpha$ by their corresponding gradients $\nabla\alpha_i = \frac{\delta \hat{y}}{\delta \alpha_i}$, where $\hat{y}$ is the predicted label, and is shown to work better than using only the attention weights; (ii) \textbf{DeepLIFT/ DL} \cite{pmlr-v70-shrikumar17a} that assigns scores based on the difference between activation of each neuron and a reference activation (zero embedding vector). Note that although \newcite{liu-avci-2019-incorporating} have used the Integrated Gradients (IG) \cite{pmlr-v70-sundararajan17a}, we use DL as it is most often a good and a faster approximation of IG \cite{DBLP:conf/iclr/AnconaCO018}.

\section{Experimental Setup}

\subsection{Data}

We use three standard hate-speech datasets, namely, \textit{Waseem} \cite{waseem}, \textit{HatEval} \cite{basile-etal-2019-semeval} and \textit{Vidgen} \cite{vidgen-etal-2021-learning}. 
Following
\citet{wiegand-etal-2019-detection,swamy-etal-2019-studying}, we perform hate\slash non-hate classification across domains.
We use the standard splits available for \textit{HatEval} (42.1\% hate; train: 8993\footnote{The instances containing only URLs are removed, decreasing the number of train instances from 9000 to 8993.}, val: 1000; test: 3000) and \textit{Vidgen} (54.4\% hate; train: 32497, val: 1016, test: 4062). We sub-sample the \textit{Vidgen} validation set by 25\% to get 1016 samples, making its size similar to the other datasets. We split \textit{Waseem} (26.8\% hate) randomly into train (80\%; 8720), validation (10\%; 1090) and test (10\%; 1090) sets, as no standard splits are available.

We present the top ten most frequent terms in these datasets in Table \ref{Top_freq}. The \textit{Waseem} dataset is known to comprise a high proportion of implicit hate \citep{wiegand-etal-2019-detection}, which are subtle expressions of hate without the use of profanity. This is also evident in the most frequent terms from this dataset. In Table \ref{Top_freq}, \textit{\#mkr} refers to a cooking show which frequently results in sexist comments targeted towards the participating women.  
\textit{HatEval} involves hate against women and immigrants. Many hateful tweets against immigrants occurred in the context of the US-Mexico border issues with the hashtag \textit{\#buildthewall}. The \textit{Vidgen} dataset is collected through a dynamic data creation process with a human-and-model-in-the-loop strategy, unlike \textit{HatEval} and \textit{Waseem} datasets that are sampled from Twitter. In particular, the \textit{Vidgen} dataset involves hate against many different target groups or identity terms, with a wide variety of topics and hateful forms.
See Appendix \ref{sec:append-Data} for further details on the datasets.

\begin{table}[!h]
\centering
\begin{tabularx}{\columnwidth}{p{1.65cm}| p{5.2cm}}
\hline \textbf{Dataset} & \textbf{Frequent terms in the datasets} \\ \hline
\textit{Waseem} & \#mkr, \#notsexist, kat, women, like, andre, get, people, one, think \\ \hline
\textit{HatEval} & b*tch, women, refugees, \#buildthewall, immigrant, immigration, illegal, men, migrants, h*e  \\ \hline
\textit{Vidgen} & people, black, women, f*cking, like, love, think, white, get, want \\
\hline
\end{tabularx}
\caption{\label{Top_freq} Top ten most frequent terms in the datasets after removing the stop-words.} 
\end{table}

\begin{table*}[!t]
\scriptsize
\centering
\begin{tabularx}{\textwidth}{p{4cm}|p{1.25cm}|p{1.25cm}|p{1.25cm}|p{1.25cm}|p{1.25cm}|p{1.25cm}|c}
\toprule
\textbf{Approaches} & \textbf{H \textrightarrow V} & \textbf{V \textrightarrow H} & \textbf{H \textrightarrow W} & \textbf{W \textrightarrow H} & \textbf{V \textrightarrow W}&\textbf{W \textrightarrow V} & \multicolumn{1}{c}{\textbf{Average}}  \\
\hline
BERT Van-MLM-FT & 56.6$\pm$1.3& 66.2$\pm$1.2& 70.0$\pm$2.5&50.9$\pm$2.1& 61.4$\pm$2.4&43.5$\pm$1.9& 58.1\\
\newcite{liu-avci-2019-incorporating} & 45.1$\pm$4.5 & 59.5$\pm$0.7 & 57.2$\pm$3.8 & 52.6$\mbox{*}\pm$0.8 &57.1$\pm$2.7 & 39.6$\pm$2.0& 51.9\\
MLM + \newcite{kennedy-etal-2020-contextualizing} (a) & 55.4$\pm$2.0 & 65.5$\pm$0.8 & 64.1$\pm$1.4 & \underline{54.4}$\mbox{*}\pm$1.3  & 59.2$\pm$1.8 & 44.5$\pm$2.9 & 57.2\\
MLM + \newcite{kennedy-etal-2020-contextualizing} (b) & 54.9$\pm$2.9 & 65.7$\pm$0.9 & 67.3$\pm$1.2 & 54.3$\mbox{*}\pm$2.2 & 62.3$\pm$2.7 & 46.6$\pm$3.5 & 58.5 \\
BERT PERL& 54.1$\pm$0.7 &60.0$\pm$0.6 &60.1$\pm$2.0 &\textbf{55.2}$\mbox{*}\pm$0.7 &55.5$\pm$1.0 &37.8$\pm$1.2 & 53.8 \\

BERT-AAD& 56.6$\pm$1.3 & 53.9$\pm$3.5 & 68.8$\pm$2.5 &50.7$\pm$1.4& 48.3$\pm$4.7 &\textbf{53.0}$\mbox{*}\pm$1.7& 55.2\\

HATN & 48.4$\pm$1.6 & 59.1$\pm$0.4 & 59.7$\pm$2.9 & 51.4$\pm$1.8 & 60.0$\pm$2.6 & 45.4$\pm$2.7 & 54.0\\
MLM + \newcite{Sarwar2022UnsupervisedDA} & 55.0$\pm$1.9 & 66.2$\pm$2.0 & 68.8$\pm$1.1 & 48.2$\pm$3.1 & 57.9$\pm$1.3 & 36.2$\pm$1.1  & 55.4\\
MLM + \citet{attanasio-2022-entropy} & 54.9$\pm$1.6 & 66.5$\pm$1.4 & 64.1$\pm$5.0& 52.4$\mbox{*}\pm$3.7 & 62.5$\pm$0.8 & 43.5$\pm$2.3& 57.3\\
MLM + ${\chi}^{2}$-test & 57.9$\pm$1.6 & 67.1$\pm$1.7 & 69.8$\pm$0.8 & 48.2$\pm$3.1 & 60.4$\pm$2.8 & 44.1$\pm$3.4 & 57.9 \\\hline
Pre-def ($\alpha \nabla \alpha$) & \textbf{58.9}$\mbox{*}\pm$0.7 & \underline{67.4}$\pm$1.5 & \underline{71.3}$\pm$1.0 & 48.9$\pm$4.0 & 60.0$\pm$2.0 & 46.5$\pm$4.9 & 58.8 \\
Dom-spec ($\alpha \nabla \alpha$) & 58.3$\pm$1.8 & 66.8$\pm$0.7 & 70.1$\pm$1.8 & 52.3$\mbox{*}\pm$3.0 & 60.8$\pm$2.2 & 46.9$\mbox{*}\pm$2.5 & 59.2\\
Comb (Dom-spec + Pre-def) ($\alpha \nabla \alpha$) & 58.7$\mbox{*}\pm$2.1 & \textbf{67.7}$\pm$1.0 & 70.9$\pm$1.0 & 51.5$\pm$2.1 & 59.8$\pm$1.5 & 45.9$\pm$3.1  & 59.1 \\ \hline
Pre-def (DL) & 58.5$\mbox{*}\pm$1.4 & 66.5$\pm$1.3 & 70.3$\pm$1.7 & 51.2$\pm$1.7 & \textbf{70.3}$\mbox{*}\pm$0.5 & 42.7$\pm$2.0 & 59.9\\
Dom-spec (DL) & \underline{58.8}$\mbox{*}\pm$0.6 & 66.4$\pm$1.2 & \textbf{72.2}$\pm$1.4 & 52.9$\mbox{*}\pm$1.9 
& 63.6$\mbox{*}\pm$2.0 & \underline{48.8}$\mbox{*}\pm$4.7 & \underline{60.5}\\
Comb (Dom-spec + Pre-def) (DL) & 58.4$\pm$1.4 & 66.7$\pm$1.0 & \underline{71.3}$\pm$0.9 & 51.1$\pm$2.2  
& \underline{69.5}$\mbox{*}\pm$2.2 & 46.6$\pm$1.9 & \textbf{60.6} \\
\bottomrule
\end{tabularx}
\caption{\label{Results_DA}
Macro-F1 ($\pm$std-dev) on source \textrightarrow target pairs. H: HatEval, V: Vidgen, W: Waseem. 
\textbf{Bold} denotes the best score and \underline{underline} the second best in each column . $\mbox{*}$ denotes statistically significant improvement compared to Van-MLM-FT with paired bootstrap \protect\citep{dror-etal-2018-hitchhikers, efron1994introduction}, 95\% confidence interval.}
\end{table*}

\subsection{Baselines}
We compare our work with approaches that penalize \textbf{(i)} pre-defined terms in Convolutional Neural Networks-based \citet{liu-avci-2019-incorporating}\footnote{with Integrated Gradients \cite{pmlr-v70-sundararajan17a}}; \textbf{(ii)} (a) the identity terms in the top features of a bag-of-words Logistic Regression in BERT-based \citet{kennedy-etal-2020-contextualizing}\footnote{with Sampling and Occlusion \cite{Jin2020Towards}} (b) all the terms listed by \citet{kennedy-etal-2020-contextualizing}; \textbf{(iii)} terms extracted automatically by \citet{attanasio-2022-entropy}; \textbf{(iv)} combination of terms from (i) and (ii,b) within BERT, and call this \textbf{Pre-Def}. We do not compare with \citet{D-Ref} as they use labeled target instances for term-extraction, which does not allow a fair comparison.

Further, we experiment with 
the Vanilla baseline (\textbf{Van-MLM-FT}), where the pre-trained BERT is adapted to $D_T^{train}$ using the MLM objective, followed by a supervised fine-tuning on $D_S^{train}$.  
We also assess different DA methods from the sentiment classification task, namely, \textbf{BERT PERL} (Pivot-based Encoder Representation of Language) \cite{ben-david-etal-2020-perl} that adopts the MLM objective of BERT to perform pivot-based fine-tuning; \textbf{BERT-AAD} (Adversarial Adaptation with Distillation) \cite{Ryu2020KnowledgeDF} that 
performs domain adversarial training; \textbf{HATN} (Hierarchical Attention Transfer Network) \cite{li2018hatn,li2017end} that extracts pivots using a domain adversarial approach. 

We evaluate a data-augmentation-based  approach \cite{Sarwar2022UnsupervisedDA} for DA in hate-speech. 
For a fair comparison, 
we use the BERT as the underlying model in this approach.
Finally, we apply the \textbf{$\bm{\chi^2}$-test} with 1 degree of freedom and Yate's correction \citep{jbp:/content/journals/10.1075/ijcl.6.1.05kil}, penalizing the terms from $D_S^{train}$, using their DL scores, for which the null hypothesis of both $D_S^{train}$ and $D_T^{train}$ being random samples of the same larger population, is rejected with 95\% confidence. 
We initialize all the BERT models with MLM adaptation on the target, except for PERL and AAD, which inherently adapts to the target.

\subsection{Model training}

We train all the models on $D_S^{train}$, use a small amount of the labeled $D_T^{val}$ 
only for model-selection and hyper-parameter tuning (see Appendix \ref{sec:append-hyper-param}) , following \newcite{dai2020adversarial,maharana-bansal-2020-adversarial}, and evaluate on $D_T^{test}$.

\section{Results}
\subsection{Discussion}

Table \ref{Results_DA} displays the macro-F1 scores obtained, in cross-domain settings,
averaged across five randomly initialized runs.
We use macro-F1 as penalizing $\text{te}^S$ 
corrects the mis-classifications for both the 
hate and non-hate 
classes across domains.
We observe an overall performance drop, compared to Van MLM-FT, with the DA approaches, originally
proposed for sentiment classification, 
namely, BERT PERL, BERT-AAD and HATN. This also agrees with 
\citet{bose-etal-2021-unsupervised}, who 
analyze the extracted pivots -- terms that are both frequent across domains as well as important for classification with respect to the source --
and find them to be sub-optimal for DA in hate-speech. The 
approach by \citet{Sarwar2022UnsupervisedDA} also displays an overall drop.
They augment the source domain by substituting relevant terms from a different negative emotion dataset with tagged hate-speech related terms from the target domain.
We observe that the augmented instances are often incomprehensible after such substitution.

Dom-spec yields improvements over 
all the baselines using both $\alpha \nabla\alpha$ and DL, both independently and in combination (Comb) with Pre-def, where Comb achieves the highest overall performance with DL: 60.6. With DL, Dom-spec yields  significantly improved performance in 4/6 cases, compared to 2/6 with Pre-def (DL). 
This 
is apparently due to the penalization of relevant source-specific terms that have wider coverage compared to the pre-defined terms in Pre-def. Since the entropy-based attention regularization by \citet{attanasio-2022-entropy} do not use the target domain unlabeled instances for term-extraction, it may not be optimal for cross-domain settings. 
The large improvement with Pre-def (DL) for \textit{Vidgen} \textrightarrow \textit{Waseem} (70.3) could be attributed to the fact that \textit{Vidgen}  
involves a wide variety of identity terms. Thus, penalizing the 
pre-defined  
identity terms 
might result in higher emphasis on more generalizable hate-speech content. While only this particular case drives the high average performance with Pre-def (DL), Dom-spec (DL) performs well 
\textit{consistently} and yields a higher average score (Dom-spec: 60.5, Comb: 60.6) compared to Pre-def. 

As discussed by \citet{wiegand-etal-2019-detection}, the \textit{Waseem} dataset includes a high degree of implicit hate. Still, Dom-spec (DL) yields improvements on the \textit{Waseem} dataset when using it as the target domain, compared to Van MLM-FT. This is reflected in the cases of  \textit{HatEval} \textrightarrow \textit{Waseem} and  \textit{Vidgen} \textrightarrow \textit{Waseem}. This is most likely because when the source domain-specific terms causing bias are penalized, the model is forced to learn more from the wider contextual meaning of the instances, rather than focusing on individual terms. We believe that this  
could possibly help in improving the detection of implicit hate in out of domain instances, at least to some extent. We leave further investigation in this direction for future work.

\subsection{Qualitative Analysis}
\label{quality}

\begin{table}[!t]
\scriptsize
\centering
\begin{tabularx}{\columnwidth}{p{3.5cm} p{3.5cm}}
\hline \multicolumn{2}{p{7cm}}{\textbf{Non-hate example from the test set of \textit{HatEval} for \textit{Waseem} \textrightarrow \textit{HatEval}}}\\
\hline \textbf{FP with Van-MLM-FT}& \textbf{TN with Dom-spec (DL)} \\ 
\colorbox{purple!6.5}{\strut Depression} \colorbox{purple!10.332500000000001}{\strut is} \colorbox{purple!17.037499999999998}{\strut a} \colorbox{purple!12.387500000000001}{\strut whole} \colorbox{purple!26.25}{\strut entire} \colorbox{purple!22.0925}{\strut b*tch}&\colorbox{purple!40.2775}{\strut Depression} \colorbox{purple!10.462499999999999}{\strut is} \colorbox{purple!6.5}{\strut a} \colorbox{purple!14.8175}{\strut whole} \colorbox{purple!17.459999999999999}{\strut entire} \colorbox{purple!12.845}{\strut b*tch}

  \\
\hline\hline
\multicolumn{2}{p{7cm}}{\textbf{Hate example from the test set of \textit{Waseem} for
\textit{Vidgen} \textrightarrow \textit{Waseem}}}\\
\hline
\textbf{FN with Van-MLM-FT}& \textbf{TP with Dom-spec (DL)} \\

...\colorbox{purple!3.7840448726365254}{\strut good} \colorbox{purple!4.2938604252563035}{\strut to} \colorbox{purple!3.246716608642153}{\strut talk} \colorbox{purple!5.251881344140691}{\strut with} \colorbox{purple!5.269526830373199}{\strut your} \colorbox{purple!2.2169292022134983}{\strut wife} \colorbox{purple!11.139388386596629}{\strut but} \colorbox{purple!3.947978702728717}{\strut it} \colorbox{purple!4.67564563825295}{\strut is} 
\colorbox{purple!2.9330869037702203}{\strut easier} \colorbox{purple!5.647442713119003}{\strut to} \colorbox{purple!8.7027954731983}{\strut say} \colorbox{purple!3.3074196058220156}{\strut shut} \colorbox{purple!3.3369865197821804}{\strut up} \colorbox{purple!12.490783663333396}{\strut n} \colorbox{purple!2.2838369716795595}{\strut make} \colorbox{purple!3.44304752967348}{\strut me} \colorbox{purple!2.715364395852024}{\strut a} \colorbox{purple!15.507096312347615}{\strut sammich} \colorbox{purple!4.959677882012582}{\strut not} \colorbox{purple!19.012699290712806}{\strut sexist} \colorbox{purple!6.020059029425544}{\strut lol}

 &

...\colorbox{purple!2.9566797013503043}{\strut good} \colorbox{purple!3.08015149355739}{\strut to} \colorbox{purple!2.634096080752119}{\strut talk} \colorbox{purple!2.56967705477614}{\strut with} \colorbox{purple!1.953125}{\strut your} \colorbox{purple!18.443507326128861}{\strut wife} \colorbox{purple!2.0782927850929185}{\strut but} \colorbox{purple!2.9351668768603196}{\strut it} \colorbox{purple!3.1527046433287054}{\strut is}
\colorbox{purple!3.0369479230275926}{\strut easier} \colorbox{purple!2.755965805646203}{\strut to} \colorbox{purple!3.0147749111189075}{\strut say} \colorbox{purple!15.2493779010952375}{\strut shut} \colorbox{purple!3.1487703495208343}{\strut up} \colorbox{purple!2.8359055687020076}{\strut n} \colorbox{purple!6.636417394547612}{\strut make} \colorbox{purple!12.480428916653904}{\strut me} \colorbox{purple!29.966588265522507}{\strut a} \colorbox{purple!35.79314910523797}{\strut sammich} \colorbox{purple!20.749717237950467}{\strut not} \colorbox{purple!19.288859983942686}{\strut sexist} \colorbox{purple!2.763012792154432}{\strut lol}

\\

\hline
\end{tabularx}
\caption{\label{qualit-analysis} Change in attributions with Dom-spec (DL).} 
\end{table}

Table \ref{qualit-analysis} displays examples of False Positives (FP) for  \textit{Waseem} \textrightarrow \textit{HatEval} and False Negatives (FN) for \textit{Vidgen} \textrightarrow \textit{Waseem}, yielded by Van-MLM-FT for the respective target domain instances, which are correctly classified by Dom-spec (DL), where the hate class is the positive class. The darker the shades, the higher the attributions assigned by the source classifier. The examples suggest that penalizing source-specific terms results in placing more emphasis on the general contextual meaning of the out-of-domain instances such as `depression' in the first example and `wife...shut...make me a sammich' in the second.

Note that the terms in these examples from the target domain that receive reduced importance with Dom-spec, compared to Van-MLM-FT, may not be the same terms that are extracted and penalized. This is because the domain classification step results in obtaining terms that are more likely to be infrequent in the target domain. Rather, due to the penalization of source-specific terms, the source domain classifier learns to focus on the wider context of the instances. For example, we observe that in the case of \textit{Waseem} \textrightarrow \textit{HatEval}, the automatically extracted $\text{te}^S$ includes terms related to the role of women in sports, 
such as \{\textit{sports, sexist, gaming, football, commentary, competition, ...}\}.  Note that  \citet{wiegand-etal-2019-detection} also mention that these terms cause domain or topic bias in \textit{Waseem}, restricting generalizability. 
See Appendix \ref{sec:append-tokens} for 
more examples.

\section{Conclusion}

We proposed a DA approach for automatic extraction and penalization of source domain-specific terms that have higher attributions towards the hate-speech labels, to improve cross-domain hate-speech detection. The results demonstrated consistent improvements on the target domain. These results should motivate further research on domain adaptation in hate-speech and building classifiers that can generalize well to the concept of hate. Finally, it would be interesting in applying our approach to other tasks such as rumor and misinformation detection~\citep{10.7717/peerj-cs.325,10.1145/3501247.3531559}. 

\section*{Ethical Considerations}
This work serves as a means to build more robust hate-speech detection models that can make proper use of the existing curated hate-speech resources and adapt well on new resources or social-media comments, which have not been well-annotated due to time and cost constraints. The hate-speech resources used for the work are publicly available and cited appropriately, wherein the authors have discussed the sampling techniques and annotation guidelines in detail. The hate-speech examples presented in the paper are only intended for research purposes and better analysis of the models explored. The terms extracted and penalized in this work are not meant to be used off-the-shelf, but the approach should serve as a starting point for research on model-debugging and building more generalizable hate-speech classifiers. 

\section*{Acknowledgments}

This work was supported partly by the french PIA project ``Lorraine Université d'Excellence'', reference ANR-15-IDEX-04-LUE. Experiments presented in this article were carried out using the
Grid'5000 testbed, supported by a scientific interest group hosted by Inria and including CNRS, RENATER and several Universities as well as other organizations (see \url{https://www.grid5000.fr}). We thank the anonymous reviewers for their valuable feedback and suggestions.

\bibliography{acl_latex}
\bibliographystyle{acl_natbib}

\appendix

\section{Differences across Datasets}
\label{sec:append-Data}

The datasets \textit{HatEval} \citep{basile-etal-2019-semeval} and \textit{Waseem} \citep{waseem} have been sampled from Twitter. \textit{HatEval} has primarily been collected in the year 2018 using a combination of sampling strategies, including keyword-based sampling (with both neutral and derogatory words), collecting the history of identified perpetrators and monitoring the potential victims of hate. It mainly consists of hate against women and immigrants. In the case of \textit{Waseem}, tweets are collected particularly using keyword-based sampling in or before 2016, with keywords that are likely to co-occur with hateful content. \citet{wiegand-etal-2019-detection} discuss the presence of a large amount of topic-bias in the dataset \textit{Waseem}. Since this dataset is available as tweet-IDs, we observe that in the crawled dataset, many tweets flagged as racist are missing, and have most likely been deleted already. Thus, the majority of available hateful content in this dataset is directed against women. The topics discussed in these two datasets are also quite different.

\textit{Vidgen}\footnote{We use an older version of the dataset. The authors have uploaded a newer version of this dataset currently.} \citep{vidgen-etal-2021-learning}, on the other hand, is a dataset generated using a human and model-in-loop process. This process results in adding several perturbations and instances, which are difficult to classify, aimed at making the dataset robust. 
Besides, it consists of hateful content directed against a wide array of target groups, e.g. \textit{black}, \textit{gay}, \textit{muslim}, \textit{disabled}, etc., along with different forms of hate such as \textit{derogation}, \textit{threatening language}, \textit{animosity}, \textit{support for hateful entities} and \textit{dehumanization}. Thus there is a substantial amount of differences across these datasets in terms of collection time-frames, sampling strategies, targets of hate, forms of hate, vocabulary used and the like.

For pre-processing the datasets, we remove the URLs, split the hashtags  
using CrazyTokenizer\footnote{\url{https://redditscore.readthedocs.io}} and lowercase the terms.

\section{Implementation Details and Hyper-parameter Tuning}
\label{sec:append-hyper-param}

We use the pre-trained BERT-base \citep{devlin-etal-2019-bert} uncased model\footnote{\url{https://huggingface.co/bert-base-uncased}} \citep{wolf-etal-2020-transformers} for our experiments. We run both the Masked Language Model (MLM) training on the unlabeled target domain training data $D_T^{train}$, and the subsequent supervised fine-tuning on the source domain training data $D_S^{train}$ for 6 epochs with a batch size of 8 for all the BERT baselines and Dom-spec. We use the AdamW optimizer with decoupled weight decay regularization \citep{Loshchilov2019DecoupledWD}, having a weight decay of $10^{-4}$. We use a learning rate of $3 \times 10^{-5}$ for the MLM training and $1 \times 10^{-5}$ for the supervised fine-tuning, with the epsilon parameter set to $1 \times 10^{-8}$. 

We use the original implementations provided by the respective authors of all the baselines except for \citet{Sarwar2022UnsupervisedDA}. We implement the data-augmentation approach by \citet{Sarwar2022UnsupervisedDA} ourselves, as there is no available implementation. We follow the description provided in the paper and label all the terms in the hateful instances of the source domain that have a match with hatebase.org\footnote{\url{https://hatebase.org/}} for training a sequence tagger. However, while finding the matches, we do not tokenize the multi-word phrases in hatebase.org. We lowercase the terms from hatebase.org and look for an exact match of a term in the source domain.

For Pre-Def, we combine the curated list of identity terms provided by \citet{liu-avci-2019-incorporating} and \citet{kennedy-etal-2020-contextualizing} and penalize their attribution scores. 
We perform hyper-parameter tuning and model selection with early-stopping on a small amount of labeled target domain validation set $D_T^{val}$ using the macro-F1 score for the proposed approach as well as for all the baselines. The hyper-parameter $\lambda$, both for the proposed approach and Pre-Def, is selected from the range $\lambda \in$ \{0.01, 0.05, 0.1, 1.0, 10.0, 20.0, 30.0, 40.0, 50.0, 60.0\}, using a random seed by tuning over $D_T^{val}$. We set the value of $M$ to 250 and $N$ to 750 for all our experiments. 

\begin{table*}[!htb]
\small
\centering
\begin{tabularx}{\textwidth}{|p{2.9cm}|p{1.4cm}|p{1.4cm}|p{1.4cm}|p{1.4cm}|p{1.5cm}|p{1.5cm}|p{1.0cm}|}
\hline
\multicolumn{1}{|l|}{\textbf{Approaches}} & \multicolumn{2}{c|}{\textbf{HatEval}} & \multicolumn{2}{c|}{\textbf{Vidgen}} & \multicolumn{2}{c|}{\textbf{Waseem}} & \textbf{Mean}  \\
\hline
BERT Van-FT & \multicolumn{2}{c|}{43.3$\pm$1.8} & \multicolumn{2}{c|}{85.1$\pm$0.5} & \multicolumn{2}{c|}{85.4$\pm$0.7} & 71.3 \\
\hline
\multicolumn{8}{|c|}{\textbf{Performance on source domain (left of arrows) while applying domain adaptation for the target (right of arrows)}}\\
\hline
 & \textbf{H \textrightarrow V} & \textbf{H \textrightarrow W} & \textbf{V \textrightarrow H}  & \textbf{V \textrightarrow W} & \textbf{W \textrightarrow H} &\textbf{W \textrightarrow V} &   \\
\hline
Dom-spec ($\alpha \nabla \alpha$) & 42.4$\pm$2.5 & 42.0$\pm$4.1 & 84.0$\pm$0.9 & 84.5$\pm$1.0 & 85.1$\pm$0.7 & 83.8$\pm$0.8 & 70.3 \\

Dom-spec (DL) & 41.7$\pm$3.7 & 40.5$\pm$4.4 & 83.9$\pm$0.7 & 82.6$\pm$1.5 & 84.7$\pm$1.2 & 81.1$\pm$2.7 & 69.1 \\
\hline
\end{tabularx}
\caption{\label{In-corpus} Effect of domain adaptation for the target on the source domain performance;
Source-domain macro average F1 scores (mean$\pm$std-dev) are obtained after MLM training on the unlabeled target domain and penalizing the source specific terms while adapting the model to the target domain (present at the right hand side of the arrows) using Dom-spec. H : HatEval, V : Vidgen, W : Waseem. Van-FT: BERT model evaluated in-domain \textit{without} MLM training on the target domain.}
\end{table*}

\section{Terms Extracted}
\label{sec:append-tokens}

The full-list of penalized terms (BERT WordPieces) $\text{te}^S$ across epochs for the examples listed in Section \ref{quality}, is given below.
\newline
\textbf{Waseem \textrightarrow HatEval}
\begin{itemize}
    \item \textbf{Epoch 1:}
\{college, sports, feminism, la, magnetic, used, unique, \#\#ava, speech, \#\#js, tr, \#\#cking, object, chu, result, ki, bus, \#\#is, adopt, referring, \#\#roids, handed, \#\#em, sh, \#\#omp, unconscious, anger, gamer, prove, xbox, tri, skill, judgment, tool, block, single, harassment, size, georgia, involved, \#\#ism, studying, voices, possible, gaming, pl, \#\#il, helped, \#\#ke, survey, equality\}
    \item \textbf{Epoch 2:}
\{feminism, used, football, awesome, equal, \#\#cking, object, \#\#ification, interest, feminist, \#\#rra, scientist, \#\#al, ignorance, bodies, \#\#work, later, \#\#nk, troll, \#\#ss, based, adopt, \#\#cing, quality, sister, unconscious, criticisms, pro, notch, xbox, tri, unfair, rap, meanwhile, impression, single, harassment, bonus, georgia, constant, sex, \#\#ist, possible, click, competition, \#\#per, swedish, \#\#eral, november, write, eventually, equality\}
    \item \textbf{Epoch 3:}
\{sham, anger, pull, used, focus, speech, ashley, object, interest, bringing, \#\#na, eye, \#\#nk, later, quality, \#\#roids, oppressive, rain, \#\#omp, statistics, nsw, content, notch, museum, unconscious, typically, tri, \#\#ol, unfair, writing, \#\#chan, georgia, constant, annie, ra, weights, click, \#\#il, furniture, helped, shopping, football, commentary, equality\}
    \item \textbf{Epoch 4:}
\{minded, kat, used, equal, focus, \#\#hand, tr, \#\#cking, chu, interest, bringing, thor, fm, \#\#tag, path, scientist, precious, later, mike, quality, humanist, \#\#roids, \#\#el, \#\#omp, worth, unconscious, nsw, xbox, tri, unfair, nu, kaitlyn, \#\#ering, pest, fe, camera, giant, constant, weights, gaming, rap, \#\#il, swedish, opposes, \#\#thi, november, laughing, survey, equality\}
    \item \textbf{Epoch 5:}
\{feminism, raging, equal, focus, \#\#hand, \#\#cking, \#\#cky, \#\#tag, \#\#na, mostly, scientist, \#\#al, \#\#rra, adopt, humanist, ft, \#\#roids, \#\#el, \#\#omp, example, unconscious, museum, anger, typically, tri, unfair, impression, yu, single, fe, cu, \#\#rd, \#\#ification, constant, grass, gaming, rap, science, \#\#per, swedish, il, furniture, shopping, november, equality\}
\end{itemize}

Few of the extracted terms get repeated in subsequent epochs as a single epoch may not be sufficient to reduce the effect of a term and it may appear in the next epoch as well. Moreover, as the training progresses, the model may learn new patterns, and some extracted terms may reappear and disappear again due to the penalization.

Following is a \textit{non-hateful} example in \textit{HatEval}, wrongly classified by Van-MLM-FT but correctly classified by Dom-spec (The darker the shades, the higher the attribution scores assigned):\\

\paragraph{Van-MLM-FT:}

\colorbox{purple!2.895997382978364}{\strut Unfortunately} \colorbox{purple!5.894858706944426}{\strut you} \colorbox{purple!4.929818519312139}{\strut are} \colorbox{purple!12.407055556455932}{\strut in} \colorbox{purple!5.0240362734375035}{\strut a} \colorbox{purple!3.0910454556609532}{\strut sticky} \colorbox{purple!8.207486215104861}{\strut size} \colorbox{purple!5.471729792275706}{\strut my} \colorbox{purple!6.151769409663247}{\strut only} \colorbox{purple!6.866130013703981}{\strut problem} \colorbox{purple!5.377947736860098}{\strut is} \colorbox{purple!15.446568538522074}{\strut replacing} \colorbox{purple!3.5535120889884237}{\strut my} \colorbox{purple!2.34375}{\strut shoes} \colorbox{purple!4.635902815927872}{\strut has} \colorbox{purple!13.511635542591092}{\strut been} \colorbox{purple!6.2483949942045225}{\strut a} \colorbox{purple!48.9739937661968}{\strut b*tch}

\paragraph{Dom-spec (DL):} 

 \colorbox{purple!48.95181085537551}{\strut Unfortunately} \colorbox{purple!25.213283329716315}{\strut you} \colorbox{purple!12.679381842981094}{\strut are} \colorbox{purple!14.21311097801898}{\strut in} \colorbox{purple!6.942635774099066}{\strut a} \colorbox{purple!2.34375}{\strut sticky} \colorbox{purple!4.738338047480417}{\strut size} \colorbox{purple!7.907548822847848}{\strut my} \colorbox{purple!6.713742560696393}{\strut only} \colorbox{purple!21.654908367466266}{\strut problem} \colorbox{purple!8.592159496918304}{\strut is} \colorbox{purple!25.445545057077496}{\strut replacing} \colorbox{purple!30.787617141343315}{\strut my} \colorbox{purple!25.620727726552055}{\strut shoes} \colorbox{purple!23.928039908930828}{\strut has} \colorbox{purple!17.650396515898688}{\strut been} \colorbox{purple!7.205908327553134}{\strut a} \colorbox{purple!9.917577529949755}{\strut b*tch}
\\

\textbf{Vidgen \textrightarrow Waseem}
\begin{itemize}
    \item \textbf{Epoch 1:}
\{wheelchair, \#\#zzi, dali, seekers, \#\#oons, koreans, \#\#tos, \#\#ware, \#\#ders, handicapped, principles, mac, pregnant, \#\#tier, \#\#iers, \#\#wear, \#\#bib, barren, \#\#tite, dyke\}
    \item \textbf{Epoch 2:}
\{customer, pip, principles, \#\#tos, \#\#hon, les, ko, vietnamese, teenagers, \#\#lock, \#\#sion, \#\#has, \#\#gin, \#\#rmi, poles, buddhist, handicapped\}
    \item \textbf{Epoch 3:}
\{pak, homosexuality, koreans, pleasant, \#\#tos, mirror, spaniards, \#\#fs, ro, \#\#rmi, boom, handicapped\}
    \item \textbf{Epoch 4:}
\{\#\#cky, pak, chin, \#\#tos, bender, herr, catholics, ro, buddhist\}
   \item \textbf{Epoch 5:}
\{pip, pak, \#\#tos, yellow, bender, koreans, \#\#mit, \#\#sion, \#\#has, \#\#rk, \#\#gin, catholics, ro, arrogance\}
\end{itemize}

Following is a \textit{non-hateful} example in \textit{Waseem}, wrongly classified by Van-MLM-FT, but correctly classified by the proposed approach (The darker the shades, the higher the attribution scores assigned):

\paragraph{Van-MLM-FT:}
\colorbox{purple!18.03363874264168}{\strut Omg} \colorbox{purple!6.127853420181861}{\strut I} \colorbox{purple!5.135747334856454}{\strut am} \colorbox{purple!25.07638419953137}{\strut lisening} \colorbox{purple!8.244221350708257}{\strut to} \colorbox{purple!7.709696875918227}{\strut an} \colorbox{purple!9.173052939693003}{\strut apple} \colorbox{purple!1.1361082776782603}{\strut genius} \colorbox{purple!0.10000000000000031}{\strut dude} \colorbox{purple!8.320580445943092}{\strut tell} \colorbox{purple!36.75450529650974}{\strut this} \colorbox{purple!65.158908280536}{\strut old} \colorbox{purple!80.477678981289}{\strut woman} \colorbox{purple!40.52799992562581}{\strut how} \colorbox{purple!3.1533476400564218}{\strut to} \colorbox{purple!9.554877307535135}{\strut use} \colorbox{purple!23.982591208334274}{\strut email} \colorbox{purple!15.090441626102812}{\strut and} \colorbox{purple!18.620916404770238}{\strut it} \colorbox{purple!9.354274816364494}{\strut is} \colorbox{purple!4.337372411756923}{\strut adorable}

\paragraph{Dom-spec (DL):}
\colorbox{purple!21.391045405028702}{\strut Omg} \colorbox{purple!7.178467616661478}{\strut I} \colorbox{purple!18.46598244888185}{\strut am} \colorbox{purple!0.2509041943294874}{\strut listening} \colorbox{purple!1.6455185373055767}{\strut to} \colorbox{purple!8.166281600389432}{\strut an} \colorbox{purple!38.70933584670555}{\strut apple} \colorbox{purple!1.3546421583248027}{\strut genius} \colorbox{purple!17.283771587913623}{\strut dude} \colorbox{purple!0.10000000000000031}{\strut tell} \colorbox{purple!3.2143324641051447}{\strut this} \colorbox{purple!17.75910012353146}{\strut old} \colorbox{purple!44.27199784212858}{\strut woman} \colorbox{purple!14.44816965205792}{\strut how} \colorbox{purple!4.493455882779883}{\strut to} \colorbox{purple!0.4505751918289749}{\strut use} \colorbox{purple!0.5058217336776922}{\strut email} \colorbox{purple!3.422048077774356}{\strut and} \colorbox{purple!48.76426364698219}{\strut it} \colorbox{purple!0.3425237781354557}{\strut is} \colorbox{purple!80.477678981289}{\strut adorable}

\section{In-domain Performance}
\label{sec:append-in-domain}

Table \ref{In-corpus} presents, as a reference, the in-domain macro-F1 scores using BERT supervised fine-tuning (Van-FT) \textit{without} the MLM training on the target domain. In this case, the model is tuned over the in-domain validation set. The \textit{HatEval} dataset is part of a shared task and involves a challenging test set with low in-domain performance. Table \ref{In-corpus} displays the source-domain scores obtained when source-specific terms are penalized, while adapting to the target domain using Dom-spec, where the model is tuned over the target domain validation set. The drop in in-domain performance is expected as Dom-spec is aimed at making the model best-suited to the target domain. However, the overall performance with Dom-spec is comparable to that of BERT Van-FT.

\section{List of Identity Terms for Pre-Def}
\label{sec:group-iden}

The combined list of pre-defined curated identity terms from \citet{liu-avci-2019-incorporating} and \citet{kennedy-etal-2020-contextualizing} are given below:

\{lesbian, gay, bisexual, trans, cis, queer, lgbt, lgbtq, straight, heterosexual, male, female, nonbinary, african, african american, european, hispanic, latino, latina, latinx, canadian, american, asian, indian
middle eastern, chinese, japanese, christian, buddhist, catholic, protestant, sikh, taoist, old, older, young, younger, teenage, millenial, middle aged, elderly, blind, deaf, paralyzed, muslim, jew, jews, white, islam, blacks, muslims, women, whites, gay, black, democrat, islamic, allah, jewish, lesbian, transgender, race, brown, woman, mexican, religion, homosexual, homosexuality, africans\}

\begin{table}[!t]
\centering
\small
\begin{tabularx}{\columnwidth}{p{2.65cm}|p{1.2cm}|p{1.19cm}|p{1.25cm}}
\toprule
{\textbf{Approaches}} & \textbf{HatEval} & \textbf{Vidgen} & \textbf{Waseem} \\
\hline
{BERT Van-MLM-FT} & 1 m 20 s & 3 m 49 s & 2 m 10s\\
{Dom-spec ($\alpha\nabla\alpha$)} & 2 m 30s & 7 m & 3 m 17 s \\
{Dom-spec (DL)} &  4 m & 18 m  & 8 m 16 s\\

\hline
\end{tabularx}
\caption{\label{Computation_time}
Per epoch training time on different source domains.}
\end{table}

 \section{Computational Efficiency}

 The per-epoch training time for Dom-spec, while performing adaptation of different source domain models, are presented in Table \ref{Computation_time}. Dom-spec ($\alpha \nabla \alpha$) takes less than double the time taken by Van-MLM-FT to train, and Dom-spec (DL) takes roughly 4.5 times of the training time taken by Van-MLM-FT.

\end{document}